\documentclass[hf]{ceurart}
\usepackage{listings}
\usepackage[utf8]{inputenc}
\usepackage{tikz}
\usetikzlibrary{shapes,backgrounds,svg.path}
\usepackage{verbatim}
\usepackage{url}
\usepackage{enumitem}
\usepackage{amsmath,amssymb,amsfonts}
\usepackage{xfrac}
\usepackage{algorithmic}
\usepackage{graphicx}
\usepackage{graphics}
\usepackage{svg}
\usepackage{textcomp}
\usepackage{xcolor}
\usepackage{color,soul}
\usepackage{scalerel}
\usepackage{csquotes}
\usepackage{subcaption}
\usepackage{multirow}
\usepackage[ruled,vlined]{algorithm2e}
\usepackage{nomencl}
\usepackage{comment}
\usepackage{float}

\newcommand{\citeauthornum}[1]{\citeauthor{#1} (\citeyear{#1}) \cite{#1}}

\usepackage{color}
\newcommand{\pinaforecomment}[4]{\colorbox{#1}{\textcolor{#4}{\parbox{.8\linewidth}{#2: #3}}}}
\newcommand{\osullikomment}[1]{\pinaforecomment{green}{Kent}{#1}{black}}

\begin{document}

\copyrightyear{2024}
\copyrightclause{Copyright for this paper by its authors.
  Use permitted under Creative Commons License Attribution 4.0
  International (CC BY 4.0).}

\conference{LNSAI 2024: First International Workshop on Logical Foundations of Neuro-Symbolic AI, August 05, 2024, Jeju, South Korea};

\title{Neuro-Symbolic AI in 2024: A Systematic Review}

\author[1]{Brandon C. Colelough}[%
orcid=0000-0001-8389-3403,
email=brandcol@umd.edu,
url=https://brandoncolelough.com/,
]
\cormark[1]
\fnmark[1]
\address[1]{University of Maryland, College Park,
  8125 Paint Branch Dr, College Park, MD 20742}

\author[2]{William Regli}[%
orcid=0000-0001-7116-9338,
email=regli@umd.edu,
url=https://isr.umd.edu/clark/faculty/902/William-Regli,
]

\cortext[1]{Corresponding author.}
\fntext[1]{These authors contributed equally.}

\begin{abstract}
\textbf{Background:} The field of Artificial Intelligence has undergone cyclical periods of growth and decline, known as AI summers and winters. Currently, we are in the third AI summer, characterized by significant advancements and commercialization, particularly in the integration of Symbolic AI and Sub-Symbolic AI, leading to the emergence of neuro-symbolic AI.

\textbf{Objective:} This paper provides a systematic literature review of neuro-symbolic AI projects within the 2024 AI landscape, highlighting key developments, methodologies, and applications. It aims to identify where quality efforts are focused in 2024 and pinpoint existing research gaps in the field.

\textbf{Methods:} The review followed the PRISMA methodology, utilizing databases such as IEEE Explore, Google Scholar, arXiv, ACM, and SpringerLink. The inclusion criteria targeted peer-reviewed papers published between 2020 and 2024. Papers were screened for relevance to neuro-symbolic AI, with further inclusion based on the availability of associated codebases to ensure reproducibility.

\textbf{Results:} From an initial pool of 1,428 papers, 167 met the inclusion criteria and were analyzed in detail. The majority of research efforts are concentrated in the areas of learning and inference (63\%), logic and reasoning (35\%), and knowledge representation (44\%). Explainability and trustworthiness are less represented (28\%), with meta-level cognition being the least explored area (5\%). The review identifies significant interdisciplinary opportunities, particularly in integrating explainability and trustworthiness with other research areas.

\textbf{Discussion:} The findings reveal a well-integrated body of work in learning and inference, logic and reasoning, and knowledge representation. However, there is a notable gap in research focused on explainability and trustworthiness, which is critical for the deployment of reliable AI systems. The sparse representation of meta-level cognition highlights the need for further research to develop frameworks that enable AI systems to self-monitor, evaluate, and adjust their processes, enhancing autonomy and adaptability.

\textbf{Conclusion:} Neuro-symbolic AI research has seen rapid growth since 2020, with concentrated efforts in learning and inference. Significant gaps remain in explainability, trustworthiness, and meta-level cognition. Addressing these gaps through interdisciplinary research will be crucial for advancing the field towards more intelligent, reliable, and context-aware AI systems.

\osullikomment{You can drop the 'contributed equally' from after your name if you're the only one, and I think you should use Bill's CS address: https://www.cs.umd.edu/people/regli. Does the conference specify you need to use the subheadings? If not I don't think they add much? PS I broke the bolding on 'background' and have no idea why. }
\end{abstract}


\maketitle

\section{Introduction} \label{sec:intro}
The field of Artificial Intelligence (AI) has experienced significant cyclical growth, known as AI summers and winters. At present, we as a community find ourselves in the third AI summer, marked by rapid scientific advances and commercialization, continuing the legacy of previous periods of AI excitement followed by setbacks \cite{Kautz2022}. A significant product of the third AI summer has been the integration of two prominent fields of AI; Symbolic AI and Sub-Symbolic AI, the fusion of which is known as neuro-symbolic AI. 
\osullikomment{Caps of Symbolic and Sub-Symbolic, but not neurosymbolic is odd.}
There is an ongoing debate about the necessity of neuro-symbolic AI \cite{Dingli2023}, with critics arguing that the world can be solved through big data and big computing and proponents arguing that \enquote{You can’t get to the moon by climbing successively taller trees} \cite{Marcus2019}. 
\osullikomment{'world can be solved' is very handwavy if not a direct quote, might be worth being more precise}
For this systematic review, we take the stance that symbolic AI is essential and that neuro-symbolic AI represents the best way forward for the community hence, this paper provides a systematic literature review of prominent neuro-symbolic projects within the 2024 AI landscape, highlighting key developments, methodologies, and applications. There has been significant interest recently in the field of neuro-symbolic AI. 
\osullikomment{Last sentence isn't a good link sentence and reads awkwardly - I think just delete it. Instead signpost what the rest of this section will be about}

\subsubsection{\textbf{Symbolic AI}}\label{subsub:symbolic-AI}
As described by \citeauthornum{Dingli2023}, Symbolic AI is a \enquote{a sub-field of AI concerned with learning the internal symbolic representations of the world around it} where we can \enquote{ translate some form of implicit human knowledge into a more formalized and declarative form based on rules and logic}. Examples of some of the earliest AI systems that utilised symbolic representations include SHRDLU from \citeauthornum{SHRDLU}, ELIZA from \citeauthornum{Weizenbaum1966},  DENDRAL from \citeauthornum{Lindsay1980} and MYCIN from \citeauthornum{Melle1978} and examples of some of the newest AI systems which heavily utilise symbolic processes include ConceptNet 5.5 by \citeauthornum{Speer2016} CYC from \citeauthornum{Lenat2023} and Good Old Fashioned AI (GOFAI) planning systems as described by \citeauthornum{Edelkamp2004} to name just a few.
\osullikomment{Having the IEEE and Author Name reads weirdly. You could reclaim a half paragraph here just listing the systems without including the author names.}

\subsubsection{\textbf{Sub-Symbolic AI}}
By contrast, Sub-symbolic-AI, as again described by \citeauthornum{Dingli2023} are systems that \enquote{do not require rules or symbolic representations as inputs} and instead \enquote{learn implicit data representations on their own}. Sub-symbolic AI encompasses approaches such as machine learning, deep learning, and generative AI etc., which rely on algorithms to automatically extract patterns from raw data to discern relationships and make predictions based on learned representations. 
\osullikomment{Definition is a bit awkward, you can drop the etc here and let 'encompasses' do some heavy lifting. If you can pull a formal definition from somewhere it'll add more weight. I'll see if I can dig one up.}
Examples of some of the earliest AI systems that utilised sub-symbolic representations include the Perceptron developed by \citeauthornum{Rosenblatt1958}, Hopfield Networks developed by \citeauthornum{Hopfield1982} and the Backpropagation Algorithm Formulated by \citeauthornum{Rumelhart1986} and examples of some of the newest sub-symbolic systems include famous projects such as the Generative Pre-trained Transformer (GPT) models by \citeauthornum{Vaswani2017}, the YOLO family of Convolutional Neural Networks (CNN'S) by \citeauthornum{Redmon2015} and the DALLE diffusion model transformer from \citeauthornum{Ramesh2021} to again just name a few. 
\osullikomment{Jumping from general architectures like perceptron to specific implementations like YOLO is a bit jarring. I think sticking to a more general level might be more consistent here e.g. Use `Attention is All You Need' instead of GPT, Yann Lecunn's CNN paper instead of YOLO etc.}
Figure \ref{fig:sub-and-symbolic-AI} below shows the difference between symbolic and sub-symbolic AI and how symbolic representations can emerge from sub-symbolic processes to be subsequently mapped to a symbolic feature map.
\osullikomment{Broken Figure Reference, I assume you're removing it. If you do include it, I think you're better off highlighting the difference in text rather than saying the figure will show it to me, don't make the reader do extra work.}

\osullikomment{Similar to last para, you can save a bunch of space just saying the architecture/algo names and not including the Author's names. Having their names adds a little but I think you could use the space better. }

\subsubsection{\textbf{Daniel Kahneman with System 1 and System 2 thinking}}
As explored by \citeauthornum{Garcez2023}, there is at present a debate within the AI community surrounding the need for neuro-symbolic AI. Simply described, the argument for neuro-symbolic AI draws on \citeauthornum{Kahneman2011} concepts of System 1 and System 2 thinking whereby System 1 is fast, intuitive, and parallel, akin to the capabilities of deep learning, while System 2 is slow, deliberate, and sequential, resembling symbolic reasoning and hence, neuro-Symbolic AI aims to combine these two approaches to create systems that benefit from the strengths of both. 
\osullikomment{I think you could merge this with the next paragraph and save a bit of space. It's a bit jarring to jump out of tech into philosophy back into tech. }

\subsubsection{\textbf{Neuro-symbolic AI}}
To clearly define exactly what we mean by the term neuro-symbolic AI, 
\osullikomment{Tautology, can replace the entire first sentence with `we adopt', which will flow smoother if you merge into back half of above.}
we rely on the definition provided by \citeauthornum{Garcez2023}; Hence, Neuro-symbolic AI is \textbf{\enquote{a composite AI framework that seeks to merge the domains of Symbolic AI and NNs} [or more broadly put, Sub-Symbolic AI] \enquote{to create a superior hybrid AI model possessing reasoning capabilities}}. As this definition is quite broad, for the purpose of this systematic review, we will further define the sub-components of the Neuro-Symbolic AI Spectrum we believe to be most relevant to the current AI landscape within section \ref{sec:methodology}. 
\osullikomment{Have you defined Neural Network yet? The `Acronym' package can be helpful for tracking that kind of thing.}

\osullikomment{You're missing the paper signposting here to explain what I'm going to read. Just needs to be a few sentences so if I'm skimming it I will see what each section will cover and skip to the interesting ones. E.g. "We provide an overview of our methodology in \autoref{sec:methodology} and present the results in \autoref{sec:results} ... }

\section{Methodology}\label{sec:methodology}

\osullikomment{Broadly agree with Bill here about this not being relevant to a conference paper. This section should be 2 paragraphs, one that defines and justifies your taxonomy/spectrum of terms (and explain why you see it as a spectrum, are the boundaries fuzzy), perhaps with a table or a nice clean visual to summarize them with a few examples and a second that says something like "We follow the \textit{PRISMA} methodology for a Systematic Literature Review for neuro-symbolic AI to extend prior survey works~\cite{Gibaut2023,Yu2021,Wan2024,Marra2024,MichelDeletie2024,Bouneffouf2022}, searching five databases to identify X candidate papers which we refine to Y papers on the basis of quality, relevance and publication language~\footnote{The detailed description of our methodology is available in the full-length version of this survey paper at: \url{http://arxiv.org}} Should allow you to reclaim at least a page of real estate}

A manual search process was conducted to identify published peer-reviewed articles. This process followed the \href{https://www.prisma-statement.org/}{PRISMA} systematic review methodology and utilised the following databases; 1. \href{https://ieeexplore.ieee.org/Xplore/home.jsp}{IEEE Explore} 2. \href{http://scholar.google.com}{Google Scholar}, 3. \href{https://arxiv.org/}{Ar\textnormal{\raisebox{0.5ex}{$\chi$}}iV Online Library}, 4. \href{http://portal.acm.org/portal.cfm}{The Association for Computing Machinery (ACM)}, 5. \href{http://www.springerlink.com}{SpringerLink Library}.

First, an initial search was conducted to identify relevant systematic reviews and survey papers for the field of neuro-symbolic artificial intelligence
\osullikomment{Inconsistent use of Caps and Abbreviation in neuro-symbolic AI in the doc}
through querying the paper title information. Only peer-reviewed papers published between 2022 and 2024 were considered. The results were further screened to remove duplicates and papers that were not relevant to the broad application space of neurosymbolic AI. The following 6 survey papers on the field of neuro-symbolic AI were then selected; \citeauthornum{Gibaut2023},  \citeauthornum{Yu2021}, \citeauthornum{Wan2024}, \citeauthornum{Marra2024}, \citeauthornum{MichelDeletie2024}, \citeauthornum{Bouneffouf2022}. 
\osullikomment{Numbers below Ten should be in words not digits unless there's a consistency reason}
The 6 selected survey papers were then used in conjunction with the following 4 books by \citeauthornum{Dingli2023}, \citeauthornum{Hitzler2023}, \citeauthornum{Hitzler2021} and \citeauthornum{Shakarian2023} to synthesize the following research question, search criteria and accept/reject criteria for the systematic literature review: 

\subsection{Research question}\label{subsec:research_quest}
\textbf{"Within the major sub-fields of neuro-symbolic AI, where is the quality of effort being focused in 2024, and what are the existing research gaps in the current literature?"}

\osullikomment{What does `Quality of Effort' mean, and is it only limited to 2024? Do you have papers from earlier than that?}

\subsection{The Neuro-Symbolic AI Spectrum}

\osullikomment{Spectrum or Taxonomy?}

\osullikomment{`To answer the above' is a weak opening, treat it like a topic sentence eg `We identify five foundational research areas advancing the state of the art for neuro-symbolic AI'}
To answer the above question, we first identified from the 6 survey papers and 4 books, the major sub-fields of neuro-symbolic AI. 5 foundational research areas were thus distilled from these 10 pieces of literature. The foundational research areas for neuro-symbolic AI, synthesized from the work of various researchers, include: 
\begin{enumerate}
    \item \textbf{Knowledge Representation}, which involves integrating symbolic and neural representations as well as developing commonsense and domain-specific knowledge graphs \cite{Gibaut2023, Hitzler2023, Shakarian2023};
    \item \textbf{Learning and Inference}, focusing on combining learning and reasoning processes through end-to-end differentiable reasoning and dynamic multi-source knowledge reasoning. May also involve structured tasks such as planning \cite{Yu2021, Wan2024, Dingli2023};
    \item \textbf{Explainability and Trustworthiness}, which aims to create interpretable models and reasoning processes to ensure trust and reliability in neuro-symbolic systems \cite{Marra2024, MichelDeletie2024};
    \item \textbf{Logic and Reasoning}, integrating logic-based methods with neural networks, including logical and probabilistic reasoning and the syntax and semantics of neuro-symbolic systems \cite{Marra2024, Shakarian2023};
    \osullikomment{How are you deliniating between KR and Logic and Reasoning. Which one would get Ontological Inference for example? }
    \item \textbf{Meta-level Cognition} involves the system's capacity to monitor, evaluate, and adjust its own reasoning and learning processes by integrating neural networks and symbolic representations.
    \osullikomment{Cite for Meta-Level Cognition? If you're introducing it, make that clearer.}

\end{enumerate}

The above categories represent the core technical areas where current efforts are concentrated from the literature skim of survey papers and currently available books on neurosymbolic AI. 
\osullikomment{Next sentence is weak, replace with a more confident statement that makes your contribution clearer "We define \textit{meta-cognition} to address a gap in current taxonomies that fail to capture <list of things> that are increasingly critical to neuro-symbolic AI research}
There is, however, an extra category, meta-level cognition, that we are adding that was not present in the literature as we believe it will be important for the field going forward. 

\subsection{PRISMA Search Criteria}

The keywords used for the search included "Neurosymbolic" combined with terms such as "Knowledge Representation," "Reasoning learning," "inference," "explainability," "trustworthiness," "architectures," "logic," "formal methods," "applications," and "challenges." The publication years were limited to 2022-2024. 
\osullikomment{Inconsistent Capitalization, and this search term violates your current research question.}
The search focused on document types such as peer-reviewed journal articles, conference papers, survey papers, systematic reviews, and books. Only articles written in English were considered. The search area was restricted to the title only, except for Springer, where the abstract was also considered.

\subsection{Accept/Reject Criteria}

The inclusion criteria comprised papers published between 2020 and 2024, articles addressing any of the seven foundational research areas, and sources that were peer-reviewed and academic. The exclusion criteria included papers published before 2020, papers not in English, survey or systematic review papers, non-peer-reviewed articles such as editorials, opinions, or preprints that were not peer-reviewed, articles not addressing the specified sub-fields of neuro-symbolic AI, duplicate entries, and any paper (EXCEPT for the meta-level cognition papers) that did not have associated publicly available codebase (e.g. GitHub, Huggingface etc), indicating non-reproducible research. 

\subsection{Screening Methodologies}
A broad search was conducted using the identified keywords across selected databases to query the literature title (title and abstract for Springer database as the title alone gave no results), and all search results were imported into JabRef. The number of records identified from each database was documented as. Duplicate records were then identified and removed using JabRef. Titles and abstracts of the collected papers were screened to assess relevance based on the inclusion and exclusion criteria, categorizing papers into "Include," "Exclude," or "Uncertain." Full texts of papers categorized as "Include" or "Uncertain" were retrieved and reviewed, rigorously applying the inclusion and exclusion criteria to determine the final list of papers to be included in the review. Data were extracted from the selected papers, including authors and publication year, title and source, major research areas addressed, and key findings relevant to the research question. The extracted data were organized into the five foundational research areas, identifying and summarizing the quality of effort and existing research gaps within each area.

\section{Results}\label{sec:results}

\osullikomment{Fig 1 is a little gross, if you're making in python I recommend you use plotnine and the theme\_classic(). See: \url{https://www.darkhorseanalytics.com/blog/data-looks-better-naked} You can ditch the title when you have a caption. Caption should probably say `histogram of' rather than `number of' and Y axis label could be more descriptive `Count of Papers Published Per Year'}

\begin{figure}[ht]
    \centering
   \includesvg[width=\textwidth]{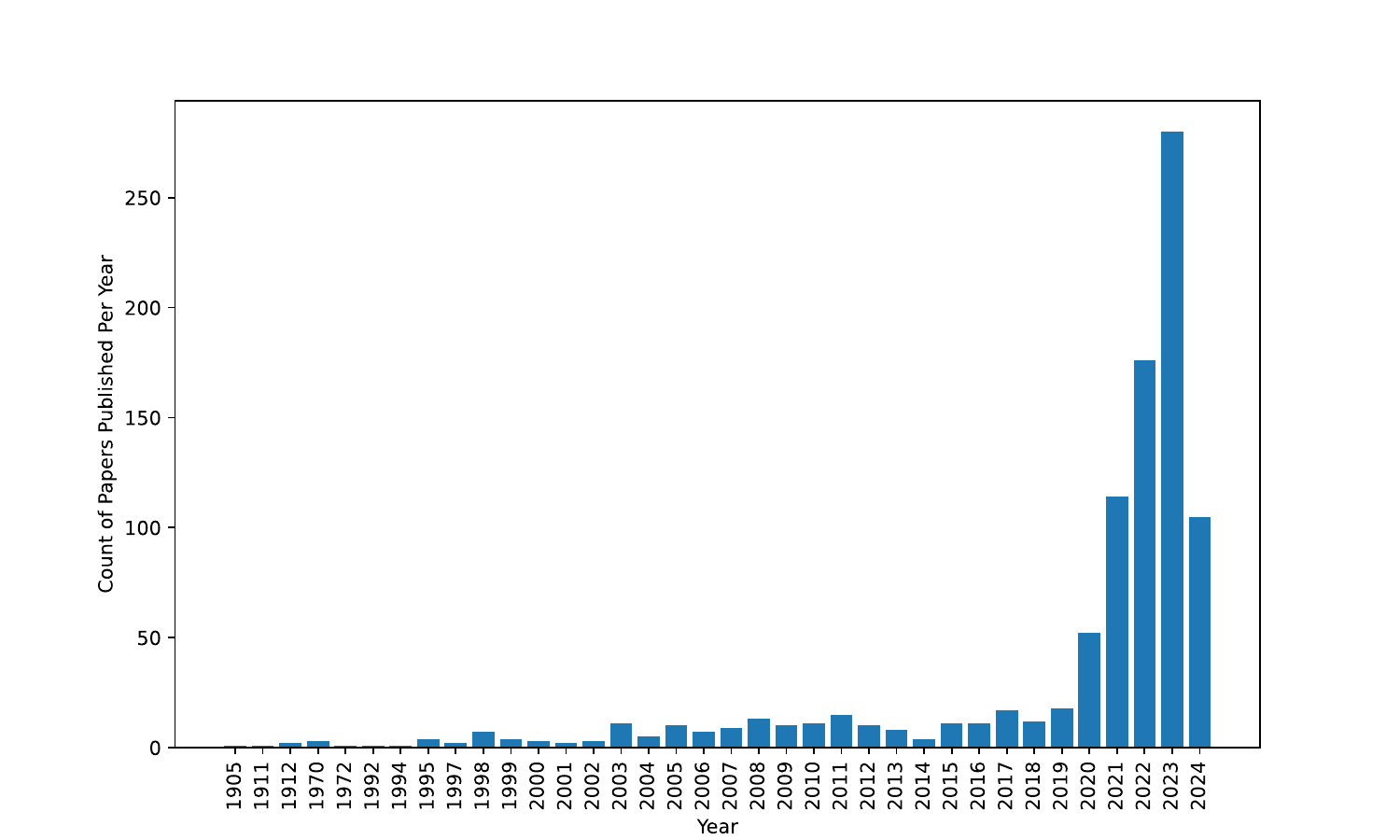}
    \caption{Number of publications per year for neuro-symbolic AI. The data was obtained through Google Scholar scraping, reflecting significant growth from 2020}
    \label{fig:pubs_by_year}
\end{figure}

\begin{figure*}[ht]
    \centering
    \resizebox{\textwidth}{!}{
    \begin{tabular}{|p{3cm}||p{3cm}|p{3cm}|p{3cm}|p{3cm}|p{3cm}|}
        \hline
        \multicolumn{6}{|c|}{Article Findings for Neuro-symbolic AI within the 5 foundational areas of research} \\
        \hline
        Database & Knowledge & Learning, Inference & Explainability, Trustworthiness & Logic, Reasoning & Meta-Level, Cognition \\
        \hline
        IEEE & 73 & 97 & 15 & 67 & 33  \\
        Google Scholar & 56 & 126 & 7 & 129 & 3  \\
        ArXiv & 17 & 54 & 7 & 55 & 3  \\
        ACM & 10 & 46 & 5 & 12 & 17  \\
        Springer & 152 & 170 & 65 & 162 & 47  \\
        Total(after screening)   & 308 & 493 & 99 & 425 & 103  \\
        \hline
     \end{tabular}

    }

\captionof{table}[Hits_Results]{The search terms "neurosymbolic" AND each of the terms required for the 5 foundational research areas within neurosymbolic AI were queried through the 5 databases. The number of pieces of literature returned from each query is shown in the table above. Note also that only publications from 2020-2024 were considered}
\label{tab::hits_2_results}
\end{figure*}

\osullikomment{What does Table1 add to your argument? If you want to keep it and you're not required to make it this format by the venue consider a nicer formatting like I've demo'd below (also - Table float environment typically better than Figure float for tables:}

\begin{table*}[ht]
    \centering
    \begin{tabular}{lrrrrr}
        Database & Knowledge & Learning \& & Explainability \& & Logic \& & Meta-Level \\
         &  & Inference & Trustworthiness & Reasoning & Cognition \\
        \hline
        IEEE & 73 & 97 & 15 & 67 & 33  \\
        Google Scholar & 56 & 126 & 7 & 129 & 3  \\
        Ar\textnormal{\raisebox{0.5ex}{$\chi$}}iV & 17 & 54 & 7 & 55 & 3  \\
        ACM & 10 & 46 & 5 & 12 & 17  \\
        Springer & 152 & 170 & 65 & 162 & 47  \\
        Total(after screening)   & 308 & 493 & 99 & 425 & 103  \\
     \end{tabular}
\end{table*}

\osullikomment{I actually really like the idea of the venn diagram to explain this, just need to (a) put the whole thing in a single figure and (b) put a bit of effort in to  improving the readability, eg: 1. Make all font sizes the same, 2. Change font color to be white for your blue and purple. Also need to figure out how to represent meta-cognition more naturally - does it really have no overlap?}

\osullikomment{Hmm, maybe a Hive Plot or an UpSet Plot will make it prettier to look at but are probably harder to interpret}

\def\firstcircle{(-1.5,1.5) circle (3cm)}
\def\secondcircle{(1.5,1.5cm) circle (3cm)}
\def\thirdcircle{(1.5,-1.5cm) circle (3cm)}
\def\fourthcircle{(-1.5,-1.5cm) circle (3cm)}

\begin{figure*}[p]
    \centering
    \resizebox{\textwidth}{!}{

    \begin{tikzpicture}

        \begin{scope}[shift={(-4cm,0cm)}, fill opacity=0.5]
            \fill[red] \firstcircle;
            \fill[green] \secondcircle;
            \fill[blue] \thirdcircle;
            \fill[yellow] \fourthcircle;
            \draw \firstcircle node[above] {};
            \draw \secondcircle node [above] {};
            \draw \thirdcircle node [above] {};
            \draw \fourthcircle node [below] {};
        \end{scope}
    \node at (-7,3) {\textbf{Logic and }};
    \node at (-7,2.5) {\textbf{Reasoning }};
    
    \node at (-1.5,3) {\textbf{Learning and }};
    \node at (-1.5,2.5) {\textbf{Inference}};
    
    \node at (-6.5,-3) {\textbf{Explainability \& }};
    \node at (-6.3,-3.5) {\textbf{Trustworthiness}};

    \node at (-1.5,-3) {\textbf{Knowledge}};
    \node at (-1.7,-3.5) {\textbf{Representation}};

    \tiny

        
        \node at (-4, 0) {\textbf{\cite{Trinh2024}}};

        \node at (-8,2) {\textbf{\cite{Fang2024}}};
        \node at (-7.5,2) {\textbf{\cite{Lima2023}}};
        \node at (-7,2) {\textbf{\cite{Lin2022}}};
        \node at (-6.5,2) {\textbf{\cite{Marron2023}}};
        \node at (-6,2) {\textbf{\cite{Saad2021}}};
        \node at (-8, 1.5) {\textbf{\cite{Shindo2021}}};
        \node at (-7.5, 1.5) {\textbf{\cite{Qian2021}}};
        \node at (-7, 1.5) {\textbf{\cite{Ciatto2021}}};


\node at (-3, 1) {\textbf{\scalebox{0.5}{\cite{Ahmed2023b}}}};

\node at (-3, 1.25) {\textbf{\scalebox{0.5}{\cite{Giri2021}}}}; 
\node at (-2.75, 1.25) {\textbf{\scalebox{0.5}{\cite{Kapanipathi2020}}}}; 
\node at (-3, 0.75) {\textbf{\scalebox{0.5}{\cite{Shah2022}}}}; 
\node at (-3.25, 0.75) {\textbf{\scalebox{0.5}{\cite{Wang2022b}}}}; 

\node at (-6.5, 0.25) {\textbf{\cite{Arabshahi2021}}};
\node at (-7, 0.25) {\textbf{\cite{Chanin2023}}};

\node at (-7.5, 0) {\textbf{\cite{Ismayilzada2023}}};
\node at (-7, 0) {\textbf{\cite{Olausson2023}}};
\node at (-6.5, 0) {\textbf{\cite{Prentzas2021}}};
\node at (-6, 0) {\textbf{\cite{Saparov2022}}};
\node at (-5.5, 0) {\textbf{\cite{Shakarian2023}}};

\node at (-6.5, -0.25) {\textbf{\cite{Yang2022}}};
\node at (-7, -0.25) {\textbf{\cite{Yang2023}}};



\node at (-4.75, 2.25) {\textbf{\scalebox{0.6}{\cite{Akl2023}}}};
\node at (-4.5, 2.25) {\textbf{\scalebox{0.6}{\cite{Bosselut2020}}}};
\node at (-4.25, 2.25) {\textbf{\scalebox{0.6}{\cite{Cha2020}}}};
\node at (-4, 2.25) {\textbf{\scalebox{0.6}{\cite{Chaudhury2021}}}};
\node at (-3.75, 2.25) {\textbf{\scalebox{0.6}{\cite{Chen2023b}}}};
\node at (-3.5, 2.25) {\textbf{\scalebox{0.6}{\cite{Chitnis2022}}}};
\node at (-3.25, 2.25) {\textbf{\scalebox{0.6}{\cite{Chitnis2022a}}}};
\node at (-3, 2.25) {\textbf{\scalebox{0.6}{\cite{Feng2022}}}};

\node at (-5, 2) {\textbf{\scalebox{0.6}{\cite{Gupta2023a}}}};
\node at (-4.75, 2) {\textbf{\scalebox{0.6}{\cite{Hazra2023}}}};
\node at (-4.5, 2) {\textbf{\scalebox{0.6}{\cite{Hazra2023a}}}};
\node at (-4.25, 2) {\textbf{\scalebox{0.6}{\cite{He2021}}}};
\node at (-4, 2) {\textbf{\scalebox{0.6}{\cite{Hsu2023}}}};
\node at (-3.75, 2) {\textbf{\scalebox{0.6}{\cite{Kimura2021}}}};
\node at (-3.5, 2) {\textbf{\scalebox{0.6}{\cite{Kimura2021a}}}};
\node at (-3.25, 2) {\textbf{\scalebox{0.6}{\cite{Krieken2022}}}};
\node at (-3, 2) {\textbf{\scalebox{0.6}{\cite{Mghames2023}}}};
\node at (-2.75, 2) {\textbf{\scalebox{0.6}{\cite{Murugesan2020}}}};

\node at (-5, 1.75) {\textbf{\scalebox{0.6}{\cite{Nguyen2022}}}};
\node at (-4.75, 1.75) {\textbf{\scalebox{0.6}{\cite{Sanders2024}}}};
\node at (-4.5, 1.75) {\textbf{\scalebox{0.6}{\cite{Sanyal2024}}}};
\node at (-4.25, 1.75) {\textbf{\scalebox{0.6}{\cite{Sharifi2023}}}};
\node at (-4, 1.75) {\textbf{\scalebox{0.6}{\cite{Silver2022}}}};
\node at (-3.75, 1.75) {\textbf{\scalebox{0.6}{\cite{Wang2022}}}};
\node at (-3.5, 1.75) {\textbf{\scalebox{0.6}{\cite{Winters2021}}}};
\node at (-3.25, 1.75) {\textbf{\scalebox{0.6}{\cite{Yan2021}}}};
\node at (-3, 1.75) {\textbf{\scalebox{0.6}{\cite{Yan2022}}}};
\node at (-2.75, 1.75) {\textbf{\scalebox{0.6}{\cite{Yan2022a}}}};

\node at (-5, 1) {\textbf{\scalebox{0.5}{\cite{Badreddine2022}}}};


\node at (-5, 0.75) {\textbf{\scalebox{0.5}{\cite{Dickens2024}}}}; 
\node at (-5.25, 1) {\textbf{\scalebox{0.5}{\cite{Kodnongbua2024}}}}; 
\node at (-4.75, 1) {\textbf{\scalebox{0.5}{\cite{Potteiger2023}}}}; 
\node at (-4.75, 0.75) {\textbf{\scalebox{0.5}{\cite{Zhang2021}}}}; 

        \node at (-2,2) {\textbf{\cite{Da2021}}};
        \node at (-0.5,2) {\textbf{\cite{Feinman2021}}};
        \node at (0,2) {\textbf{\cite{GholipourGhalandari2020}}};
        \node at (-0.5, 2.5) {\textbf{\cite{Gomaa2023}}};
        \node at (-0.5, 1.5) {\textbf{\cite{Yasunaga2022}}};
        \node at (0, 1.5) {\textbf{\cite{Zhan2021}}};
        \node at (-1, 1.5) {\textbf{\cite{Zhou2023}}};
        \node at (-1.5,1.5) {\textbf{\cite{Roberts2022}}};

        \node at (0,1) {\textbf{\cite{Fabiano2023}}};
        \node at (-0.5,1) {\textbf{\cite{Agravante2023}}};

        \node at (-1,3.5) {\textbf{\cite{Haut2022}}};
        \node at (-1.5,3.5) {\textbf{\cite{Haut2023}}};
        \node at (-2,3.5) {\textbf{\cite{Kirk2024}}};
        \node at (-0.5,3.5) {\textbf{\cite{Knowles2024}}};
        \node at (-2.5,3.5) {\textbf{\cite{Kulal2021}}};
        \node at (-3,3.5) {\textbf{\cite{Liu2022}}};

        \node at (-1.5,4) {\textbf{\cite{Ross2022}}};
        \node at (-2,4) {\textbf{\cite{Smirnova2023}}};
        \node at (-2.5,4) {\textbf{\cite{Srivastava2022}}};
        \node at (-3,4) {\textbf{\cite{Sukhbaatar2021}}};
        \node at (-3.5,4) {\textbf{\cite{Wu2022}}};

        \node at (-2, -2.5) {\textbf{\cite{Hwang2021}}};
        \node at (-1.5, -2.5) {\textbf{\cite{Ismayilzada2022}}};
        \node at (-1, -2.5) {\textbf{\cite{Mostafazadeh2020}}};
        \node at (-.5, -2.5) {\textbf{\cite{Papoulias2023}}};


        \node at (-2, -2) {\textbf{\cite{Perevalov2022}}};
        \node at (-1.5, -2) {\textbf{\cite{Ribeiro2021}}};
        \node at (-1, -2) {\textbf{\cite{Chen2023}}};

        \node at (-5, -2) {\textbf{\cite{Agafonov2022}}};
        \node at (-4.5, -2) {\textbf{\cite{Harmon2023}}};
        \node at (-4, -2) {\textbf{\cite{Racharak2021}}};
        \node at (-3.5, -2) {\textbf{\cite{Raj2023}}};
        \node at (-3, -2) {\textbf{\cite{Ribeiro2022}}};

        \node at (-8,-2) {\textbf{\cite{Chen2021}}};
        \node at (-7.5,-2) {\textbf{\cite{Dingli2023}}};
        \node at (-7,-2) {\textbf{\cite{Hessel2022}}};
        \node at (-6.5,-2) {\textbf{\cite{Kalyanpur2020}}};
        \node at (-6,-2) {\textbf{\cite{Wan2022}}};

        \node at (-8,-2.5) {\textbf{\cite{Kuennecke2024}}};


\node at (-0.5,0) {\textbf{\scalebox{0.6}{\cite{Ahmed2023}}}};
\node at (-0.8,0) {\textbf{\scalebox{0.6}{\cite{Ahmetoglu2022}}}};
\node at (-1.1,0) {\textbf{\scalebox{0.6}{\cite{Baran2022}}}};
\node at (-1.4,0) {\textbf{\scalebox{0.6}{\cite{Baugh2023}}}};
\node at (-1.7,0) {\textbf{\scalebox{0.6}{\cite{Calinescu2024}}}};
\node at (-2.0,0) {\textbf{\scalebox{0.6}{\cite{Cingillioglu2021}}}};
\node at (-2.3,0) {\textbf{\scalebox{0.6}{\cite{Cunnington2022}}}};
\node at (-2.6,0) {\textbf{\scalebox{0.6}{\cite{Dold2021}}}};

\node at (-0.75,-0.3) {\textbf{\scalebox{0.6}{\cite{Gao2022}}}};
\node at (-1.05,-0.3) {\textbf{\scalebox{0.6}{\cite{Gao2023}}}};
\node at (-1.35,-0.3) {\textbf{\scalebox{0.6}{\cite{Harmon2022}}}};
\node at (-1.65,-0.3) {\textbf{\scalebox{0.6}{\cite{He2023}}}};
\node at (-1.95,-0.3) {\textbf{\scalebox{0.6}{\cite{Hong2021}}}};
\node at (-2.25,-0.3) {\textbf{\scalebox{0.6}{\cite{Howard2023}}}};
\node at (-2.55,-0.3) {\textbf{\scalebox{0.6}{\cite{Huang2023}}}};

\node at (-1.0,-0.55) {\textbf{\scalebox{0.6}{\cite{Khatiwada2022}}}};
\node at (-1.3,-0.55) {\textbf{\scalebox{0.6}{\cite{Kim2023}}}};
\node at (-1.6,-0.55) {\textbf{\scalebox{0.6}{\cite{Klinger2023}}}};
\node at (-1.9,-0.55) {\textbf{\scalebox{0.6}{\cite{Li2020}}}};
\node at (-2.2,-0.55) {\textbf{\scalebox{0.6}{\cite{Liang2020}}}};
\node at (-2.5,-0.55) {\textbf{\scalebox{0.6}{\cite{Luo2024}}}};

\node at (-0.45,0.3) {\textbf{\scalebox{0.6}{\cite{Marconato2023}}}};
\node at (-0.75,0.3) {\textbf{\scalebox{0.6}{\cite{Marconato2023a}}}};
\node at (-1.05,0.3) {\textbf{\scalebox{0.6}{\cite{Marconato2024}}}};
\node at (-1.35,0.3) {\textbf{\scalebox{0.6}{\cite{Martone2022}}}};
\node at (-1.65,0.3) {\textbf{\scalebox{0.6}{\cite{Mukherji2024}}}};
\node at (-1.95,0.3) {\textbf{\scalebox{0.6}{\cite{Namasivayam2023}}}};
\node at (-2.25,0.3) {\textbf{\scalebox{0.6}{\cite{NunezMolina2023}}}};
\node at (-2.55,0.3) {\textbf{\scalebox{0.6}{\cite{Oruganti2024}}}};


\node at (-1.0,0.55) {\textbf{\scalebox{0.6}{\cite{Potnis2023}}}};
\node at (-1.3,0.55) {\textbf{\scalebox{0.6}{\cite{Purohit2023}}}};
\node at (-1.6,0.55) {\textbf{\scalebox{0.6}{\cite{Raymond2023}}}};
\node at (-1.9,0.55) {\textbf{\scalebox{0.6}{\cite{Ruaro2021}}}};
\node at (-2.2,0.55) {\textbf{\scalebox{0.6}{\cite{Saha2024}}}};
\node at (-2.5,0.55) {\textbf{\scalebox{0.6}{\cite{Sen2021}}}};

\node at (-0.95,0.8) {\textbf{\scalebox{0.6}{\cite{Shirai2023}}}};
\node at (-1.25,0.8) {\textbf{\scalebox{0.6}{\cite{Tao2023}}}};
\node at (-1.55,0.8) {\textbf{\scalebox{0.6}{\cite{Valenti2022}}}};
\node at (-1.85,0.8) {\textbf{\scalebox{0.6}{\cite{Wang2023}}}};
\node at (-2.15,0.8) {\textbf{\scalebox{0.6}{\cite{Werner2023}}}};
\node at (-2.45,0.8) {\textbf{\scalebox{0.6}{\cite{Wu2022a}}}};

        
\node at (-3, -1) {\textbf{\scalebox{0.5}{\cite{Ahmed2022}}}};

\node at (-3, -0.75) {\textbf{\scalebox{0.5}{\cite{Himabindu2023}}}}; 
\node at (-3.25, -0.75) {\textbf{\scalebox{0.5}{\cite{Xian2020}}}}; 
\node at (-2.75, -0.75) {\textbf{\scalebox{0.5}{\cite{Zheng2022}}}}; 
\node at (-3, -1.25) {\textbf{\scalebox{0.5}{\cite{Pinhanez2021}}}}; 
\node at (-3.25, -1.25) {\textbf{\scalebox{0.5}{\cite{Stammer2021}}}}; 
\node at (-2.75, -1.25) {\textbf{\scalebox{0.5}{\cite{Thomson2024}}}}; 
\node at (-3.25, -1) {\textbf{\scalebox{0.5}{\cite{Weir2022}}}}; 

        \normalsize
        \node at (-6,-5){\textbf{Meta-Level Cognition}};
        \tiny
        \draw (0,-5.5) -- (-8,-5.5) -- (-8,-7) -- (0,-7) -- (0,-5.5);
        
        \node at (-7.5,-6)  {\textbf{\cite{Joshi2024}}};
        \node at (-7,-6) {\textbf{\cite{Liu2024}}};
        \node at (-6.5,-6) {\textbf{\cite{McDonald2024}}};
        \node at (-6,-6) {\textbf{\cite{Raja2024}}};
        \node at (-5.5,-6) {\textbf{\cite{Harini2023}}};
        \node at (-5,-6) {\textbf{\cite{Romero2024}}};
        \node at (-4.5,-6) {\textbf{\cite{Sumers2023}}};
        \node at (-4,-6) {\textbf{\cite{West2024}}};
        
        \node at (-3.5,-6) {\textbf{\cite{Choi2024}}};

    \end{tikzpicture}
}
\caption{A literature review of existing of the major components of Symbolic AI was conducted. Note that papers from the Meta-Level Cognition were not required to have an associated public code-base/repository}
\label{fig::venn_diagram}
\end{figure*}


From the initial Google Scholar scraping, there was a total of 945 publications listed on Scholar alone from 1970 til the present, with notable increases in the years beginning from 2020 (53 publications), and peaking in 2023 (280 publications) illustrated further below in figure \ref{fig:pubs_by_year}.
\osullikomment{I think `Figure' should be capitalized in this instance. Also, consider rephrasing this to be more of a so what. e.g. "\autoref{fig:pubs_by_year} shows how research on neuro-symbolic AI is increasing exponentially starting in 2020 ..." you should be focusing on the `so what' and `therefore' rather than the `what'} 
Combining the Google Scholar literature from 2020 onwards with the literature from the 4 other databases queried for pieces of literature on Neuro-Symbolic AI from 2020 onwards, a total of 1,428 papers were extracted as illustrated in table \ref{tab::hits_2_results}. From the total literature extracted, 45\% (n = 641) were removed as duplicate entries and 28\% (n = 395) were removed during title and abstract screening with 28\% (n = 392) held for further analysis. From the remaining 392 papers, the literature was further split based on code/model availability. 42\% of the papers (n = 167) had associated code-base repositories (e.g. GitHub, Huggingface etc.) and 58\% (n = 221) were further excluded from this literature review as a public code-base could not be found for the associated piece of literature. The remaining 167 papers gathered were then read in detail and sub-categorised back under the 5 main focal research areas (Logic and Reasoning, Learning and Inference, Explainability and Trustworthiness, Knowledge Representation and Meta-Level Cognition) and the intersection found therein, with a further 9 papers removed for not meeting the inclusion criteria. An illustration of this sub-categorisation showing the overlap between 4 of the 5 main research focal areas is shown in \autoref{fig::venn_diagram}. There were 44\% (n=70) entries in the Knowledge Representation category, 63\% (n=99) entries in the Learning and Inference category, 28\% (n=44) entries in the Explainability and Trustworthiness category, 35\% (n=55) entries in the Logic and Reasoning category, 5\% (n=8) entries in the Meta-level Cognition category. The intersection of Knowledge Representation and Learning \& Inference had 27\% (n=43) entries, the intersection of Knowledge Representation and Explainability \& Trustworthiness had 4\% (n=5), the intersection of Learning \& Inference and Logic \& Reasoning had 11.48\% (n=31), the intersection of Explainability \& Trustworthiness and Logic \& Reasoning had 3.33\% (n=9). The intersection of any of the three categories had a range from 5 to 9 entries, except for the intersection of Explainability and Trustworthiness \& Logic and Reasoning \& Knowledge Representation, which had none. There was only one entry at the intersection of all 4 of the main research focal areas (excluding Meta-Level cognition) which was AlphaGeometry from Google\cite{Trinh2024}. 

\section{Discussion}\label{sec:discussion}

\osullikomment{Hmm, this needs to signpost a bit more positively, readly like you're telling me what you're not doing. Something like "We extend the discussion of prior neuro-sybmolic AI literature reviews by analyzing the ten most influential papers published in each sub-topic since 2022." ...  ideally you should tease some of your findings here to entice people to read further, I'll suggest some when I finish reading this section }

For ease of readability and since there is already a plethora of literature that summarises concisely the neuro-symbolic landscape before 2022 (see \ref{sec:methodology})
\osullikomment{either use \\autoref or manually specify `figure', `section' etc }, we choose, in this section,  to report on the state-of-the-art (SOTA) technology available for researchers. For brevity, we limit each of the sub-fields of neuro-symbolic AI to at most 10 SOTA projects. For the full list, see the Venn diagram figure \ref{fig::venn_diagram} above. 

\subsection{Knowledge Representation}\label{subsec:disc_knowledge}

\osullikomment{A bunch of these cites are pre-2022, you made a big deal about focusing on post-2022, so might need to soften that language a little before here.}

Leveraging Abstract Meaning Representation for Knowledge Base Question Answering by \citeauthornum{Hwang2021} enhances systems by providing structured semantic representations, forming a foundation for more accurate and meaningful responses. Building on the integration of structured knowledge, the Knowledge-Based Event Forecasting Toolkit presented by \citeauthornum{Ismayilzada2022} significantly improves predictive capabilities by combining event forecasting with comprehensive knowledge representation. The Commonsense Knowledge Base for Modeling Atomic Events developed by \citeauthornum{Mostafazadeh2020} advances this \osullikomment{unclear what `this' is referring to, I think you mean `advances event-based knowledge representation'}further by creating a nuanced knowledge base, enabling AI systems to understand and reason about everyday situations with greater context-awareness.
The RECKONING project by \citeauthornum{Chen2023} explores hyper-edge prediction within meta-knowledge graphs, enhancing inference capabilities by dynamically encoding complex relationships, which is crucial for sophisticated AI reasoning. To ensure consistency in AI-generated narratives, Persona Commonsense Knowledge for Consistent and Engaging Narratives by \citeauthornum{Gao2023} integrates persona-based commonsense knowledge, enhancing the engagement and coherence of storytelling agents. Similarly, the Representation of Vectors and Abstract Meanings for Information Synthesis by \citeauthornum{Perevalov2022} improves information synthesis through advanced vector and abstract meaning representations. Analyzing Commonsense Emergence in Few-shot Knowledge Models by \citeauthornum{Ribeiro2021} examines the development of commonsense knowledge in few-shot learning models, advancing the understanding of minimal training data requirements. Complementing these efforts, \citeauthornum{Ahmed2022} introduces a Unified Framework for Multi-Task Knowledge Representation Learning, which boosts efficiency across diverse tasks by integrating symbolic and sub-symbolic learning approaches. Additionally, NeuroQL: A Neuro-Symbolic Language and Dataset for Inter-Subjective Reasoning by \citeauthornum{Papoulias2023} introduces a novel task and baseline solution for combining objective and subjective information, using a neuro-symbolic language to answer inter-subjective questions. Collectively, these projects significantly advance the field by improving AI's ability to process, reason, and generate human-like understanding, essential for developing intelligent, reliable, and context-aware systems.

\osullikomment{It is not super clear to me how you've arranged these. It's not chronologically ordered based on the cites and I can't see the clear choice of ordering for how you present it. I think it will be much clearer if you synthesize the `so what' from all of these directions and then make a forecast about where you expect the work to go from here. Similarly to the intro I think you can drop all the use of author names and just focus on the outputs to get this down to a shorter length. Something like the following, (If I get it correct having not read the papers).  "Research in Knowledge Representation has focused on advancing grounding, representing complex relationships, and improving data-efficiency. The grounding of question-answering and dialog systems using knowledgebases~\cite{Hwang2021} and event-based representations~\cite{Ismayilzada2022, Mostafazadeh2020} focuses on reducing the error rate in generated text. Improving the quality of representations with complex relationships explores using hyper-edge prediction in knowledge graphs~\cite{Chen2023}, complex embeddings~\cite{Perevalov2022}, personalized knowledge representation~\cite{Gao2023} and domain specific languages~\cite{Papoulias2023} to capture increasingly complex and long-range relationships. Finally, work to improve the data efficiency in training have evaluated the emergence of commonsense reasoning in few-shot prompting contexts~\cite{Ribeiro2021} and by leveraging neuro-symbolic representations as inputs to training~\cite{Ahmed2022} continue to explore how neuro-symbolic approaches offer a mechanism to `do more with less', yielding savings in training time, data storage costs and environmental impacts. Collectively, these projects significantly advance the field by improving AI's ability to process, reason, and generate human-like understanding, essential for developing intelligent, reliable, and context-aware systems, with current trends suggesting that... <whatever you think the gap is or where best research ROI is>}

\subsection{Learning and Inference}\label{subsec:disc_learn}
The project Plan-SOFAI by \citeauthornum{Fabiano2023} introduces a neuro-symbolic architecture for AI planning inspired by Kahneman's cognitive theory, integrating fast (System-1) and slow (System-2) thinking models. This architecture demonstrates versatility and efficiency, balancing solving speed and solution optimality in various scenarios. The paper by \citeauthornum{Ahmed2023} introduces a Pseudo-Semantic Loss for Autoregressive Models, enhancing logical consistency and reducing language model toxicity across diverse tasks. In a related effort, \citeauthornum{Ahmed2023a} presents methods for Semantic Strengthening of Neuro-Symbolic Learning, improving prediction accuracy in structured-output tasks by focusing on relevant constraints. \citeauthornum{Agravante2023} introduces a neuro-symbolic framework using Logical Neural Networks for model-based reinforcement learning, showcasing its effectiveness in PDDLGym environments and the TextWorld-Commonsense domain. \citeauthornum{Kimura2021} presents a neuro-symbolic reinforcement learning approach that uses Logical Neural Networks to convert observations into logical facts, improving transparency and interpretability. \citeauthornum{Badreddine2022} integrates logic and neural networks through Logic Tensor Networks, allowing rich knowledge representation and reasoning. The work by \citeauthornum{Da2021} focuses on scalable neuro-symbolic reasoning, exploring the adaptation of commonsense knowledge models in few-shot settings. \citeauthornum{Feinman2021} develops a generative neuro-symbolic model for learning task-general representations from raw perceptual inputs, applied to the Omniglot challenge. Finally, \citeauthornum{Wu2022a} introduces ZeroC, a neuro-symbolic architecture for zero-shot concept recognition and acquisition, enhancing machines' generalization capabilities at inference time. These projects collectively advance the field by improving AI's learning and inference capabilities, essential for developing intelligent and reliable systems.

\osullikomment{Same deal with this, distil the key themes or trends and organize aroung them. Signpost the three of them, summarize the work and the so what and then link back at the end to how this theme supports neuro-symbolic AI as a whole. I think your themes from this one to coalesce around are (1) Improving ability of models to generalize Fabiano, Ahmed, Ahmed, Agravante, Feinman, Wu (2) Improve Scalability Da, (3) Improving Transparency and Explainability Kimura, Badreddine, }

\subsection{Explainability and Trustworthiness}\label{subsec:disc_xai}
\citeauthornum{IHMC_Explainable_AI} highlights the importance of transparency and interpretability in AI systems, providing an excellent ground truth resource for explainable AI (XAI). The project "Do Androids Laugh at Electric Sheep? Humor 'Understanding' Benchmarks from The New Yorker Caption Contest" by \citeauthornum{Hessel2022} introduces tasks derived from the New Yorker Cartoon Caption Contest to evaluate AI's humour understanding, emphasizing the complex relationships between images and captions. "Braid: Weaving Symbolic and Neural Knowledge into Coherent Logical Explanations" by \citeauthornum{Kalyanpur2020} presents Braid, a logical reasoner supporting probabilistic rules and custom unification functions to address the brittle matching problem in traditional reasoners. \citeauthornum{Kuennecke2024} focuses on enhancing multi-domain automatic short answer grading with a neuro-symbolic pipeline, incorporating weakly supervised annotation procedures for justification cues. The work by \citeauthornum{Chen2021}, "Structure-Aware Abstractive Conversation Summarization via Discourse and Action Graphs," proposes models for conversation summarization by incorporating discourse relations and action triples to improve summary accuracy. \citeauthornum{Kuennecke2024} explores the use of neuro-symbolic methods for explainable short answer grading, improving grading accuracy through logical reasoning and justification cue detection. 
\osullikomment{Kunnecke in here twice}
\citeauthornum{Huang2023} in "FactPEGASUS: Factuality-Aware Pre-training and Fine-tuning for Abstractive Summarization" introduces FactPEGASUS, which addresses factuality in abstractive summarization through enhanced pre-training and fine-tuning techniques. \citeauthornum{Chen2021} discusses "Structure-Aware Abstractive Conversation Summarization via Discourse and Action Graphs," focusing on improving conversation summarization through structured graphs and multi-granularity decoding. Lastly, \citeauthornum{Stammer2021} introduces a method to intervene on a Neuro-Symbolic scene representation, allowing semantic-level revisions that can identify and correct confounders, thereby enhancing trust in AI through clear and understandable decision-making.
\osullikomment{Cofounders or coreferences?}

\osullikomment{As with previous, change structure and focus on the themes and project where to from here or reflect on why focus has been in these areas. I think your themes might be: (1) Explanation Generation: Hessel, Kunnecke, Chen Yang, Huang (2) Symbolic Justification: Chen, Stammer? and (3) Sub-symbolic robustness: Kalyanpur}

\subsection{Logic and Reasoning}\label{subsec:disc_logic}
The top 10 projects for logic and reasoning within neuro-symbolic AI present significant advancements in integrating symbolic and neural methods. The paper "Getting from Generative AI to Trustworthy AI: What LLMs might learn from Cyc" by \citeauthornum{Lenat2023} discusses the limitations of current Large Language Models (LLMs) in trustworthiness and reasoning, proposing the integration of LLMs with symbolic AI systems like Cyc to enhance reliability and interpretability. "An Approach to Solving the Abstraction and Reasoning Corpus (ARC) Challenge" by \citeauthornum{Min2023} utilizes GPT-4 for solving the ARC Challenge, emphasizing the importance of self-supervised learning and multi-step reasoning in understanding complex structures. "kogito: A Commonsense Knowledge Inference Toolkit" by \citeauthornum{Ismayilzada2022} offers an innovative toolkit for generating commonsense knowledge inferences from textual input, enhancing AI's adaptability. "LinkBERT: Pretraining Language Models with Document Links" by \citeauthornum{Yasunaga2022} proposes a novel pretraining approach that incorporates document links to improve language understanding and reasoning capabilities, particularly in multi-hop reasoning tasks. "Logical Credal Network" by \citeauthornum{Qian2021} develops a framework that combines logical reasoning with probabilistic models, addressing the challenge of representing imprecise information in AI applications. "DeepStochLog: Neural Stochastic Logic Programming" by \citeauthornum{Winters2021} introduces a scalable framework that enhances traditional logic programming with neural networks, improving capabilities in complex reasoning tasks. "2P-Kt: A Logic-Based Ecosystem for Symbolic AI" by \citeauthornum{Ciatto2021} presents a comprehensive logic-based framework supporting various reasoning tasks and integrating symbolic and sub-symbolic AI. \citeauthornum{Chanin2023} introduces a framework for neuro-symbolic integration in autonomous systems, enhancing decision-making and reasoning capabilities. \citeauthornum{Huang2023} presents LASER, a hybrid neuro-symbolic system for robust logical reasoning, combining neural networks' flexibility with the precision of symbolic logic. 

\osullikomment{Same as above: Possible Themes: (1) Symbolic Augmentation of Subsymbolic Approaches: Lenat, Min, Ismaylzada, (2) Improving Training for Reasoning: Yasunaga, (3) Subsymbolic Augmenation of Symbolic Approaches: Qian, Winters, Ciatto, Chanin? Huang?}

\osullikomment{Second half of next paragraph is good, but first half is a bit slow to get there. Could probably drop everything before "...Much of the literature is cross-sectional between these four distinct areas of research..." without hurting your argument at all and getting straihg to the meat of it. I also think you should reduce the amount you talk about AlphaGeometry by about half (get rid of the narrative about what it does and focus on how it touches the research areas instead) and pull it up to go straight after "...four distinct areas of research" where you use it as an example of combining all the aspescts, THEN finish this section off with your analysis of where the gaps are, clearly stating where there is room for future work. You may even want to make a topic sentence to lead with that, but I am undecided on that. }
\subsection{Intersections of the above four research areas}
Many extraordinary projects in Neuro-Symbolic AI fall within the union of two or more of the above areas of research. For example, "Spatial-Temporal Reasoning via Probabilistic Abduction and Execution" by \citeauthornum{Zhang2021} combines spatial-temporal knowledge representation with neuro-symbolic reasoning to improve cross-configuration generalization in Raven's Progressive Matrices. "Coarse-to-Fine Neural-Symbolic Reasoning for Explainable Recommendation" by \citeauthornum{Xian2020}, integrates knowledge graphs into recommender systems, using neural symbolic reasoning to generate explainable recommendations. Outlined above is a very well-integrated body of work. Much of the literature is cross-sectional between these four distinct areas of research, however, there is a distinct lack of integration with the explainability and trustworthiness fields within the unions of the other three research areas as the density of research intersection at the unions of explainability and trustworthiness and the other three research areas is relatively sparse, indicating a significant opportunity for further interdisciplinary work in neuro-symbolic AI.
AlphaGeometry stands out as a prominent project that sits at the intersection of these four research areas: explainability \& trustworthiness, knowledge representation, Learning \& Inference and Logic \& reasoning. AlphaGeometry, as detailed by \citeauthornum{Trinh2024}, is a neuro-symbolic system designed to solve Euclidean plane geometry problems at the Olympiad level. It synthesizes millions of theorems and proofs, using a neural language model trained on large-scale synthetic data to guide a symbolic deduction engine. This integration allows AlphaGeometry to produce human-readable proofs, significantly outperforming previous methods and approaching the performance of an average International Mathematical Olympiad gold medallist. Its ability to generate and utilize synthetic data, combined with symbolic reasoning, makes AlphaGeometry a groundbreaking example of how neuro-symbolic AI can achieve advanced problem-solving capabilities, bridging gaps across multiple domains of AI research.

\subsection{Meta-level Cognition}\label{subsec:disc_meta}
\osullikomment{I think you should go through the process of defining this in what is currently your  methodology section, this feels a bit late in the game to be telling me about what it is and why I care}
Meta-level cognition refers to the processes that involve thinking about one's thinking, enabling self-awareness and self-regulation in cognitive tasks. Meta-level cognition is the controller that sits above the cognitive tasks of systems to direct energy effectively towards the correct system to handle a task. This higher-order cognition is crucial for tasks that require reflection, planning, and adaptation. Its importance lies in its ability to enhance learning, problem-solving, and decision-making, making it a key focus in neuro-symbolic AI. 
\osullikomment{Everything before here move up to the Methodology section where you first introduce it. You also need to be explicit about why the other areas do not cover meta-cognition yet and be very clear about the impact of not deliberately studying it.}
Recent projects in this domain showcase innovative approaches to integrating cognitive architectures with large language models \osullikomment{LLM acronym?} and other AI techniques to \osullikomment{are they achieving it or approximating it?} achieve advanced meta-cognitive capabilities.  \citeauthornum{Harini2023} explores the use of meta-reinforcement learning combined with logical program induction to improve financial trading strategies, highlighting the benefits of integrating symbolic features with RL algorithms. \citeauthornum{Joshi2024} and \citeauthornum{Liu2024} investigate the fusion of cognitive architectures \osullikomment{I don't think you've explained what a cognitive architecture is yet?} and LLMs, aiming to create embodied agents with enhanced general intelligence by leveraging the strengths of both approaches. \citeauthornum{McDonald2024} utilizes LLMs to convert descriptive information into dense signals for instance-based learning, improving experiential learning models. \citeauthornum{Raja2024} proposes a framework that couples cognitive reasoning with generative algorithms \osullikomment{Generative algorithms - do you mean LLMs etc here or are you talking about something else. I think `generative algorithms' is a bit broad and changes how you're talking about them} to adapt to novel conditions through conflict detection and resolution. \citeauthornum{Romero2024} and \citeauthornum{Sumers2023} present integration strategies for combining LLMs with cognitive architectures, emphasizing modular, agency, and neuro-symbolic approaches to develop more robust AI systems. These projects are somewhat rooted in the foundational principles of the Common Model of Cognition (CMC), which provides a unified framework for understanding and replicating human cognitive processes.

\osullikomment{Again, distil and group by the key themes. COuld look at (1) RL to approximate Metacognition: Harini, (2) LLMs to approximate metacognition: Joshi, Liu, McDonald, Romero, Sumers, (3) Mimicing Human Processes?: Raja? Better yet, if you can group them into 'common model of cognition' and }

\osullikomment{I vote you move both of these subsections below here (Common Model and Meta-Level) up to the methodology and then `methodology' section becomes a description of your taxonomy, and your argument for why you need to extend the way it's currently done to include meta-cognition. These just feel jammed in here, make them part of the definition or delete them. I Need to have read them to understand your analysis of Metacognition Literature above.}

\subsubsection{The Common Model of Cognition}
The Common Model of Cognition, originally called the Standard Model of the Mind by \citeauthornum{Laird2017}, offers a unified framework for human cognition. It includes perception, working memory, motor functions, declarative memory, and procedural memory, integrating cognitive architectures like ACT-R, Soar, and Sigma. \citeauthornum{Thomson2024} elaborates on integrating cognitive architectures with foundation models for cognitively guided few-shot learning, enhancing AI robustness and interpretability. \citeauthornum{West2024} discusses combining generative networks with the CMC, highlighting a neuro-symbolic approach that merges symbolic reasoning with neural networks. This method replicates human cognitive processes, advancing powerful and explainable AI systems.

\subsubsection{Meta-Level Cognition in Neuro-Symbolic AI; a new frontier}
\osullikomment{I love kahneman too, but I think you can delete all the direct quotes in here. It's good talk track for a conference but doesn't make the paper stronger, if they want to read it they can go get his book. I'd limit yourself to 3 sentences that focus on the implications for AI. }
Whilst this initial representation of neuro-symbolic AI as system 1 and system 2 level thinking is a useful tool to goal-orientate the field towards a common direction for the integration of neural and symbolic processes, the current adaptation of the human-level cognitive processing ability is too simplistic and does not yet capture the full systems-level breakdown of where the community should be investing effort to push the field forward. As Kahneman himself describes the two systems \enquote{System 1 and System 2 are so central to the story I tell in this book} (Thinking fast and slow) \enquote{that I must make it absolutely clear that they are fictitious characters. Systems 1 and 2 are not systems in the standard sense of entities with interacting aspects or parts. And there is no one part of the brain that either of the systems would call home} Kahneman furthers this point towards the end of his book stating \enquote{the two systems do not really exist in the brain or anywhere else. 'System 1 does X' is a shortcut for 'X occurs automatically.' And 'System 2 is mobilized to do Y' is a shortcut for 'arousal increases, pupils dilate, attention is focused, and activity Y is performed.'}. Human-level cognition manifests from a deeply complex and intricately layered yet densely connected box of systems of systems that work in unification and act with system-1 system-2 characteristics. The goal of the neuro-symbolic research domain is to \enquote{create a superior hybrid AI model possessing reasoning capabilities} and hence, to push the field further towards this goal we must seek to design and build systems that act with the same propensity as Kahnemans system-1 system-2 character setup through the implementation of more integrated systems of systems controlled through meta-level cognition possessing the ability to act lazily when necessary and focused when required. 

\section{Conclusion}\label{sec:conclusion}
The field of neuro-symbolic AI has experienced a notable surge in research activity from 2020 onwards, reflecting the growing recognition of the importance of integrating symbolic and sub-symbolic approaches to enhance AI's reasoning capabilities. Our systematic review highlights that the majority of high-quality research efforts in 2024 are concentrated in the areas of learning and inference, with a significant portion also dedicated to logic and reasoning, as well as knowledge representation. 
\osullikomment{Any idea why its focusing here?}
These areas have seen substantial advancements, with innovative projects and methodologies pushing the boundaries of what AI systems can achieve in terms of understanding, reasoning, and generating human-like responses. However, our review also identifies several critical gaps in the current literature. Despite the substantial progress in learning and inference, there remains a relative sparseness of research focused on explainability and trustworthiness. This gap is particularly concerning given the increasing deployment of AI systems in real-world applications, where transparency and reliability are paramount. Moreover, the intersection of the four main research areas—learning and inference, logic and reasoning, knowledge representation, and explainability and trustworthiness—reveals a significant opportunity for interdisciplinary work. The density of studies that effectively combine these domains indicates the field is generally well integrated. The most underrepresented area in our review is meta-level cognition. This emerging field, which involves systems' capacity to monitor, evaluate, and adjust their own reasoning and learning processes, holds great potential for advancing AI towards more autonomous and adaptable intelligence. The few existing studies in this domain suggest promising directions, but much more work is needed to develop robust frameworks and practical implementations of meta-level cognitive architectures.

\osullikomment{I think your conclusion is missing the 1-2 sentence summary of your contributions which to me is (1) Definition of Meta-Cognition, (2) Review of key themes in Lit 2022-2024 and (3) ID of gaps in current work and how they relate to broader needs.}

\section{Acknowledgments}

\bibliography{references}

\appendix

\end{document}